\documentclass[10pt,twocolumn,letterpaper]{article}

\usepackage{cvpr}
\usepackage{times}
\usepackage{epsfig}
\usepackage{graphicx}
\usepackage{amsmath}
\usepackage{amsmath,amssymb,amsthm}
\usepackage[nowatermark]{fixmetodonotes}
\cvprfinalcopy
\usepackage[utf8]{inputenc} 
\usepackage[T1]{fontenc}    
\usepackage{url}            
\usepackage{booktabs}       
\usepackage{amsfonts}       
\usepackage{nicefrac}       
\usepackage{microtype}      
\usepackage{enumitem}	
\usepackage[numbers,square, comma, sort&compress]{natbib}	 

\newcommand\Mark[1]{\textsuperscript{#1}}

\def \FPT{{\hyperref[FPTDef]{FPT}}}
\def \UFPT{{\hyperref[UFPTDef]{{\sc UFPT}}}}
\def \NPT{{\hyperref[NPTDef]{{\sc NPT}}}}

\usepackage[pagebackref=true,breaklinks=true,letterpaper=true,bookmarks=false]{hyperref}

\def\cvprPaperID{10439} 

\begin{document}

\title{Learnt dynamics generalizes across tasks, datasets, and populations}

\author{U. Mahmood\Mark{1}, M. M. Rahman\Mark{1}, A. Fedorov\Mark{2} , Z. Fu\Mark{1}, V. D. Calhoun\Mark{1, 2, 3}, S. M. Plis\Mark{1} \\
Tri-institutional Center for Translational Research in Neuroimaging and Data Science: \\
\Mark{1}Georgia State University, \Mark{2}Georgia Institute of Technology, \Mark{3}Emory University\\
Atlanta, GA, USA\\
{\tt\small \{umahmood1,mrahman21\}@student.gsu.edu}
{ \tt\small afedorov@gatech.edu} \\
{\tt\small \{zfu,vcalhoun,splis\}@gsu.edu} 
}

\maketitle

\begin{abstract}
  Differentiating multivariate dynamic signals is a difficult learning problem as the feature space may be large yet often only a few training examples are available.
  Traditional approaches to this problem either proceed from
  handcrafted features or require large datasets to combat the $m\gg n$ problem.
  In this paper, we show that the source of the problem---signal dynamics---can be used to our advantage and noticeably improve classification performance on a range of discrimination tasks when training data is scarce.
  We demonstrate that self-supervised pre-training guided by signal dynamics produces embedding that generalizes across tasks, datasets, data collection sites, and data distributions.
  We perform an extensive evaluation of this approach on a range of tasks including simulated data, keyword detection problem, and a range of functional neuroimaging data, where we show that a single embedding learnt on healthy subjects generalizes across a number of disorders, age groups, and datasets.
 \end{abstract}

\section{Introduction}
Many interesting domains and systems are best characterized by their mutivariate dynamics.
Short and long term weather fluctuations, speech and ambient sound, and, importantly, function of human brain to name a few.
Often in these applications we need to differentiate one condition from another based on these dynamics.
Although handcrafting features for a classifier is a common and widespread approach, discarding potentially valuable information, when we know that dynamics is the key to the system's behavior, is suboptimal.
Especially when one of the goals is to learn what determines a condition of the system.

Consider a simple problem of training devices to recognize custom keywords with minimal training. This is a setting where only a few training samples can be expected.
Arguably more important domain to deal with is the area of mental health.

Mental disorders manifest in behavior that is driven by disruptions in brain dynamics~\cite{goldberg1992common, calhoun2014chronnectome}.
Functional MRI captures the nuances of spatio-temporal dynamics that could potentially provide clues to the causes of mental disorders and enable early diagnosis.
However, the obtained data for a single subject is of high dimensionality $m$ and to be useful for learning, and statistical analysis, one needs to collect datasets with a large number of subjects $n$.
Yet, for any kind of a disorder, demographics or other types of conditions, a single study is rarely able to amass datasets large enough to go out of the $m\gg n$ mode.
Traditionally this is approached by handcrafting features~\cite{Khazaee2016} of much smaller dimension, effectively reducing $m$ via dimensionality reduction.
Often, the dynamics of brain function in these representations vanishes into proxy features such as correlation matrices of functional network connectivity (FNC)~\cite{yan2017discriminating}.
Efforts that pull together data from various studies and increase $n$ do exist, but it is difficult to generalize to study of smaller and more specific disease populations that cannot be shared to become a part of these pools or are too different from the data in them.

Our goal is to enable the direct study of systems dynamics in smaller datasets. In the case of brain data it, in turn, can enable an analysis of brain function. In this paper, we show how one can achieve significant improvement in classification directly from dynamical data on small datasets by taking advantage of publicly available large but unrelated datasets.  We demonstrate that it is possible to train a model in a self-supervised manner on dynamics of healthy control subjects from the Human Connectome Project (HCP)~\cite{van2013wu} and apply the pre-trained encoder to a completely different data collected across multiple sites from healthy controls and patients. We show that pre-training on dynamics allows the encoder to generalize across a number of datasets and a wide range of disorders: schizophrenia, autism, and Alzheimer's disease. Importantly, we show that learnt dynamics generalizes across different data distributions, as our model pre-trained on healthy adults shows improvements in children and elderly. The generality of the approach is also demonstrated in an application to the keyword detection problem.

\section{Related Work}
Unsupervised pre-training is a well-known technique to get a head start for the deep neural network. It may be considered as a regularizer which compares to classical regularizers (i.e. L1/L2) may not vanish even with more data and could find a robust local minima for better generalization~\cite{erhan2010does}. Classical methods are Deep Beliefs Networks (DBMs)~\cite{NIPS2007_3211} and stacked denoising autoencoders (SDAE)~\cite{vincent2008extracting}. Unsupervised pre-training has broad implications in fields such as computer vision~\cite{krizhevsky2012imagenet}, natural language processing (NLP) (GPT2~\cite{radford2019language}, BERT~\cite{devlin2018bert}, Word2Vec~\cite{mikolov2013distributed}) and automatic speech recognition (ASR) (with SDAE~\cite{gehring2013extracting}, with DBN-HMMs~\cite{yu2010roles}). However, this unsupervised pre-training is considered to be less popular in fields other than NLP~\cite{goodfellow2016}. Specifically, in computer vision, researchers usually use the model which is pre-trained in supervised fashion on Imagenet as a starting point for downstream tasks. Furthermore, given enough data and technical strategies, it is possible to achieve better results on COCO object detection without supervised pre-training on Imagenet~\cite{He_2019_ICCV}.

Recent advances in unsupervised learning using self-supervised methods with mutual information objectives have reduced the gap between supervised and unsupervised learning on standard computer vision classification datasets~\cite{oord2018representation, hjelm2018learning, henaff2019data, bachman2019learning, he2019momentum} and scaled pre-training to very deep convolutional networks (e.g., 50-layer ResNet). Furthermore, it influences the neuroimaging field for classification of progression to Alzheimer's disease from sMRI~\cite{fedorov2019prediction}, learning useful representation of the states from the frames in Atari games~\cite{anand2019unsupervised} and also from the speech chunks for speaker identification~\cite{ravanelli2018learning}. Specifically, authors~\cite{henaff2019data} have shown that contrastive based self-supervised pre-training can outperform supervised methods by a large margin in case of small data (e.g., $13$ images per class in ImageNet~\cite{imagenet_cvpr09}).

In most cases, due to practical reasons, researchers in brain imaging
are constrained to work with small datasets. In addition, earlier
work~\cite{khosla2019machine, frontiers2014} in brain imaging have been based on unsupervised methods to learn the dynamics and structure of the brain while supervised approaches are used to perform predictions at individual level. Such unsupervised methods include models as linear ICA~\cite{calhoun2001method}, HMM framework~\cite{eavani2013unsupervised}. Moreover, some other nonlinear approaches are also proposed to capture the dynamics. Examples include using Restricted Boltzman Machines (RBMs)~\cite{hjelm2014restricted}, RNN modification of ICA~\cite{hjelm2018spatio}, and reconstructions by recurrent U-Net architecture~\cite{khosla2019detecting}.  In some cases, where dataset is very small, transfer learning as observed in some neuroimaging application~\cite{mensch2017learning, 10.3389/fnins.2018.00491, thomas2019deep} is considered as a way to enable learning from data and thus improve results in downstream classification. To achieve improved performance, another idea is the data generating approach~\cite{ulloa2018improving} which uses synthetic data generator for pre-training, relieving the scarcity of data. 

\section{Methods}
Our method is two fold. We first pre-train our encoder on large unrelated dataset to learn improved representation of the latent factors, and then use the pre-trained encoder for downstream task. We explain both steps in the following sections. 

\subsection{Pre-training}

\subsubsection{Spatio-Temporal DeepInfoMax}
\label{STDIM}
For self-supervised pre-training we use Spatio-Temporal DeepInfoMax~\cite{anand2019unsupervised} to maximize predictability between current latent state and future spatial state and between consecutive spatial states (for example, on encoded time points of the resting-state fMRI (rsfMRI)).

Let $D = \{(u_t, v_s): 1 \le t, s \le N, t \ne s\}$ be a dataset of pairs of values at time point $t$ and $s$ sampled from sequence with length $N$. Then $D^+ = \{(u_t, v_s):  1 \le t \le N-1, s = t+1\}$ is called a dataset of positive pairs and $D^- = \{(u_t, v_s): 1 \le t, s \le N, s \ne t + 1\}$ --- of negative pairs. The dataset $D^+$ refers to a joint distribution and $D^-$ --- a marginal distribution. Eventually, the lower bound with InfoNCE estimator~\cite{infonce} $\mathcal{I}_f(D^+)$ is defined as:

\begin{equation}
 \mathcal{I}(D^+) \ge \mathcal{I}_f(D^+) \triangleq \sum^N_{t=1} \log \frac{\exp f((u_t, v_{t+1})^+)}{\sum^N_{s=1} \exp f((u_t, v_s)^-)},
\end{equation}
where $f$ is a critic function~\cite{tschannen2019mutual}. Specifically, we are using separable critic $f(u_t,v_s) = \phi(u_t)^\intercal \psi(v_s)$, where $\phi$ and $\psi$ are some embedding functions parameterized by neural networks. Such embedding functions are used to calculate value of a critic function in same dimensional space from two dimensional inputs. Critic learns an embedding function such that critic assigns higher values for positive pairs compared to negative pairs: $f(D^+) \gg f(D^-)$.

\begin{figure}[ht]
  \includegraphics[width=\linewidth]{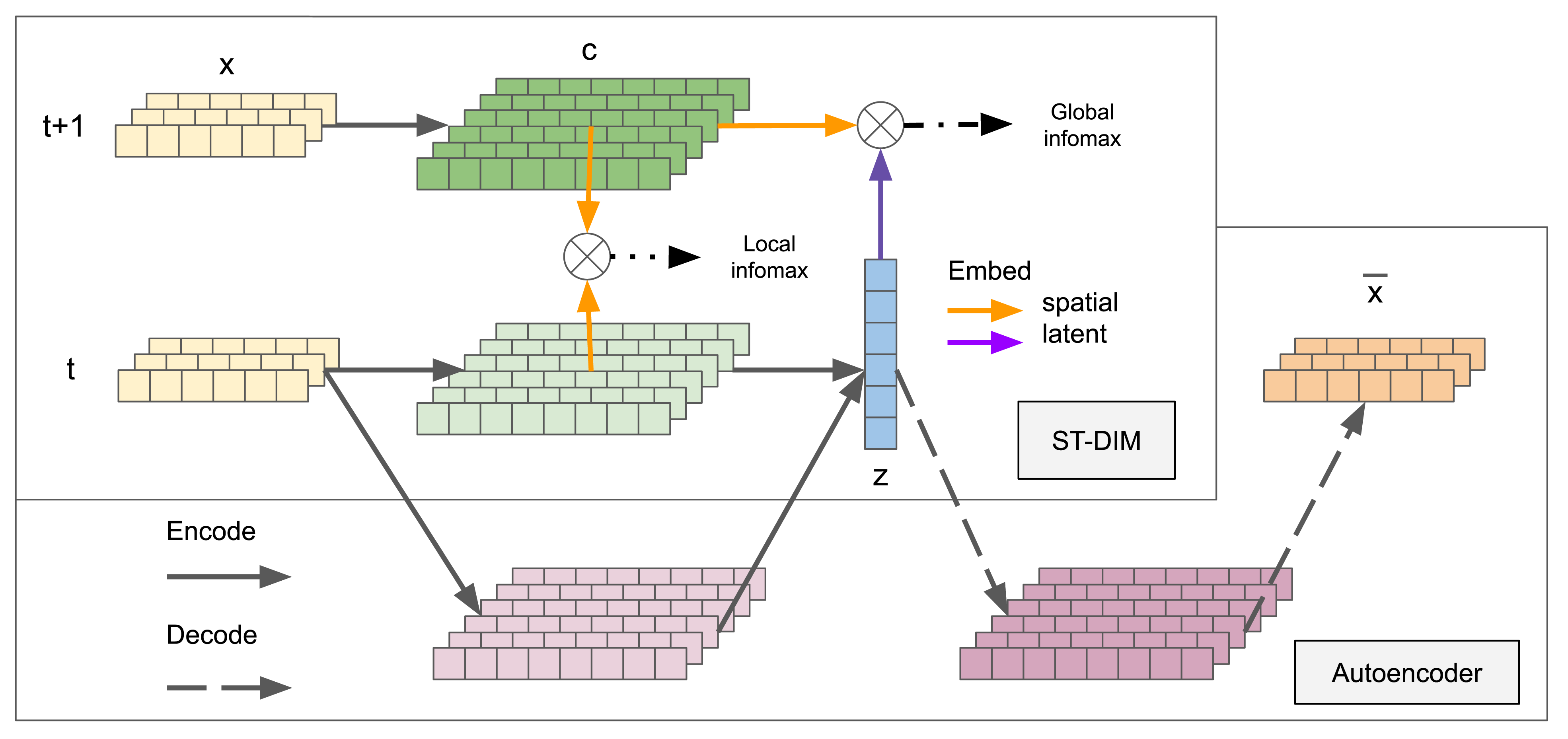}
  \centering
  \caption{Spatio-Temporal Deep InfoMax (ST-DIM) and autoencoder architecture used in pre-training. To enable comparison, the same encoder architecture is used for both ST-DIM and autoencoder models.}
  \label{fig:pre-training}
  \vspace{-1mm}
\end{figure}

We define a latent state as an output $z_t$ of encoder $E$ and a spatial state $c^l_t$ as the output of $l$th layer of the encoder for input $x_t$ at time point $t$. To optimize the objective between current latent state and future spatial state, the critic function for input pair $(x_t, x_s)$ is $f_{global} = \phi(z_t)^\intercal \psi(c^l_s)$ and between consecutive spatial states --- $f_{local} = \psi(c^l_t)^\intercal \psi(c^l_s)$. Finally, the loss is the sum of the InfoNCE with $f_{global}$ and InfoNCE with $f_{local}$ as $L = {I}_{f_{global}} + {I}_{f_{local}}$. The scheme of the ST-DIM is shown in Figure~\ref{fig:pre-training}.

\subsubsection{Autoencoder}
\label{Autoenc}
An autoencoder~\cite{autoenc1} is often viewed as an unsupervised learning representation technique useful for obtaining an informative representation of the input for downstream tasks~\cite{goodfellow2016}.
As implied in its name, it consists of two phases, namely, reduction (encoding) and reconstruction (decoding).
In the reduction phase, the autoencoder compresses its input preserving useful latent representation of the input.
In the reconstruction phase, it learns to create outputs similar to inputs based on the latent representation achieved in reduction phase.
Both phases are trained together to obtain optimal results.
The reduction and reconstruction phases can be considered as two mathematical functions $\Phi: \mathcal{X} \mapsto \mathcal{F}$ and $\Psi: \mathcal{F} \mapsto \mathcal{X}$ respectively.
We use the mean-squared error for the loss function which is defined as: $\mathcal{L}(x,\hat{x})=\|x - \hat{x}\|^2 = \|x - \Psi(\Phi(x))\|^2$, where $x, \hat{x} \in \mathcal{X}$ and refer to input and output of the autoencoder respectively.
Though autoencoder can be implemented using different types of networks, we use the same network architecture as used for ST-DIM pre-training (See Section~\ref{Setup}) to make them ideally comparable. The architecture used in pre-training is shown in Figure~\ref{fig:pre-training}.

\subsection{Transfer and Supervised Learning}
\begin{figure}[ht]
  \includegraphics[width=0.9\linewidth]{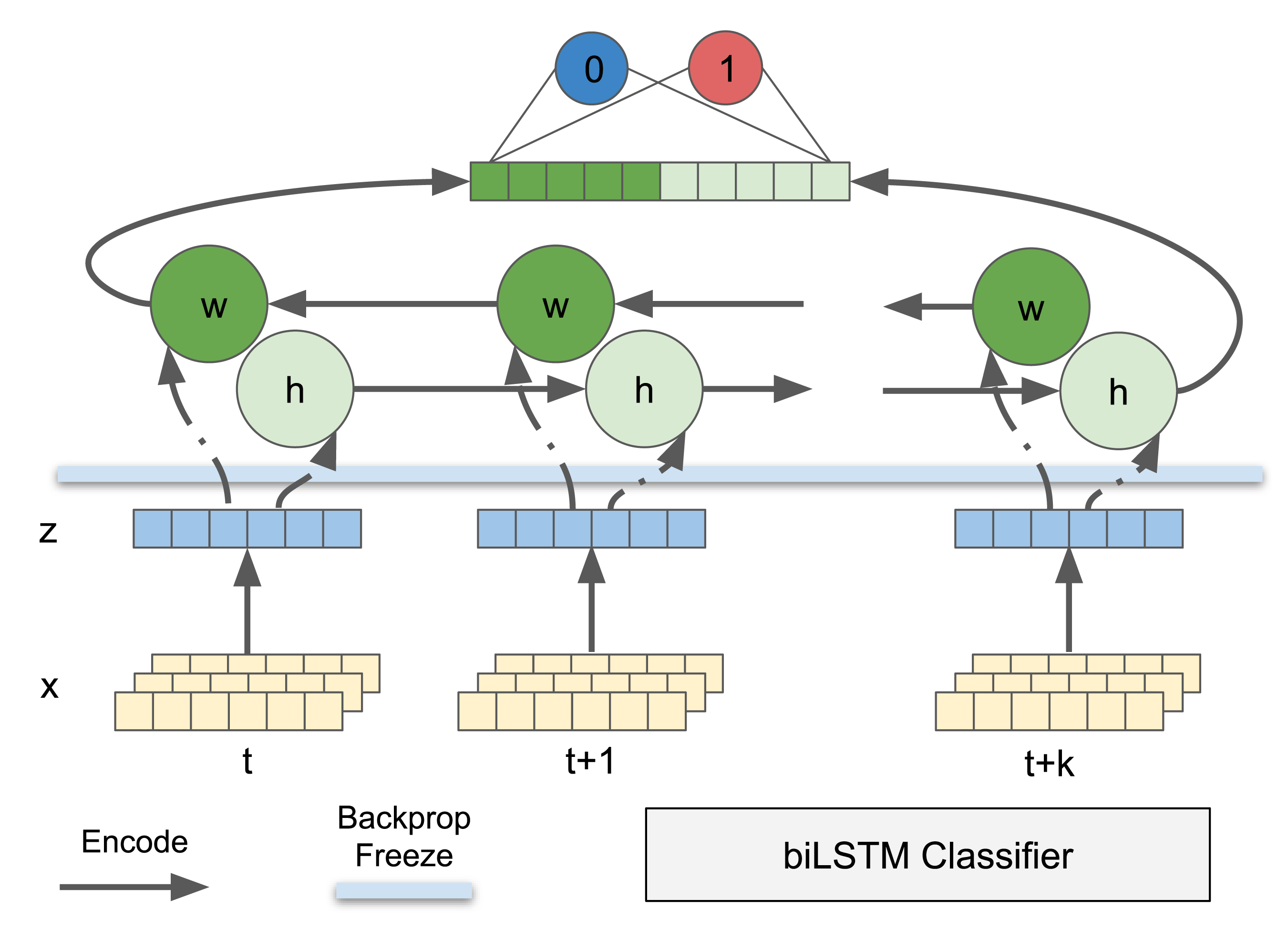}
  \centering
  \caption{Architecture used in downstream task. Windows (time slices) are encoded using encoder described in Section~\ref{Autoenc} and passed through a biLSTM. Outputs of the last forward and backward LSTM layers are concatenated and passed through a dense layer for classification.
  }
  \label{fig:model_fig}
  \vspace{-1mm}
\end{figure}

In the downstream task, we use the representation of the encoder pre-trained using ST-DIM or autoencoder as input to a biLSTM~\cite{schuster1997bidirectional} and add a simple binary classifier on top. We also use the same encoder architecture to learn end-to-end supervised model (i.e., no pre-trained weights) for the same classification tasks. The overall architecture used for downstream tasks is shown in Figure~\ref{fig:model_fig}. Refer to section~\ref{Setup} for further details. 

\section{Experiments}
\begin{figure}[ht]
  \includegraphics[width=0.9\linewidth]{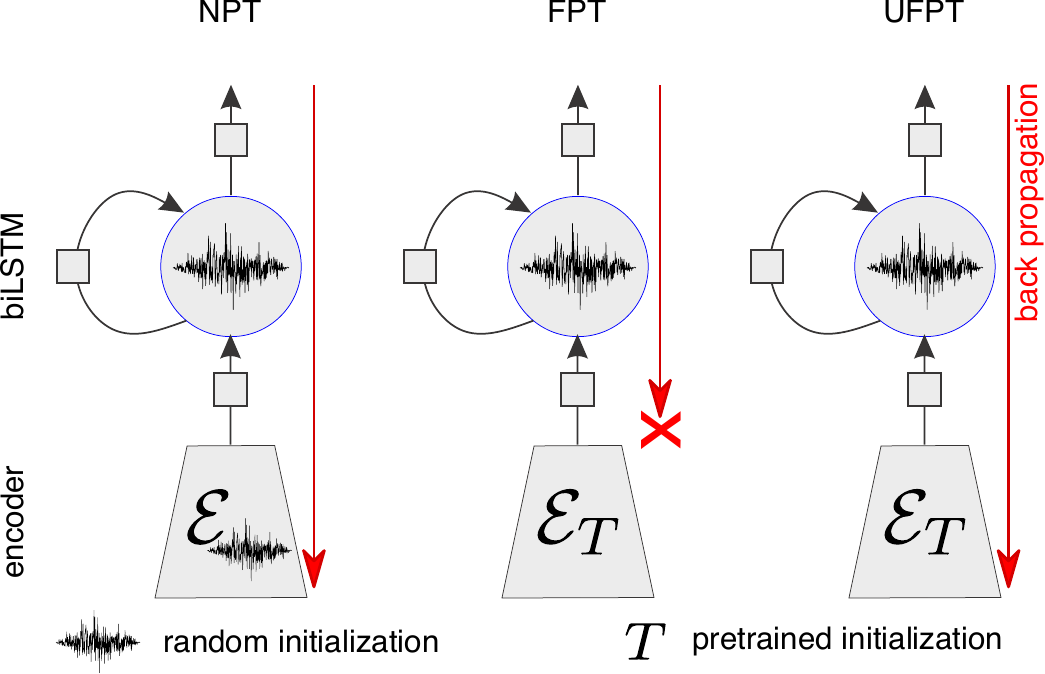}
  \centering
  \caption{Three different models are used for downstream tasks: 1) In \FPT{}, we pre-train our encoder but keep it frozen during downstream task. 2) In \UFPT{}, we pre-train our encoder but continue  training during downstream task. 3) In \NPT{}, we use the exactly same architecture but with no pre-trained weights for the encoder. 
  }
  \label{fig:process}
  \vspace{-1mm}
\end{figure}

In  this  section  we  study  the  performance  of our model on both, synthetic and real data. To compare and show the advantage of pre-training on large unrelated dataset we use three different kind of models --- 1) \phantomsection\label {FPTDef} \FPT\ (Frozen Pre-Trained): The pre-trained encoder is not further trained on the dataset of downstream task, 2)\phantomsection\label {UFPTDef} \UFPT\ (Unfrozen Pre-Trained): The pre-trained encoder is further trained on the dataset of downstream task and 3) \phantomsection\label {NPTDef} \NPT\ (Not Pre-trained): The encoder is not pre-trained at all and only trained on the small dataset of downstream task. The models are shown in Figure~\ref{fig:process}. In each experiment, we compare all three models to demonstrate the effectiveness of unsupervised pre-training.

\subsection{Setup}
\label{Setup}
Encoder for simulation experiment consists of $4$ $1$D convolutional layers with output features $(32,64,128,64)$, kernel sizes $(4,4,3,2)$ respectively, followed by ReLU~~\cite{glorot2011deep} after each layer followed by a linear layer with $256$ units. For real data experiments, we use $3$ $1$D convolutional layers with output features $(64, 128, 200)$, kernel sizes $(4, 4, 3)$ respectively, followed by ReLU after each layer followed by a linear layer with $256$ units. We use stride $1$ for all of the convolution layers.
For the autoencoder based pre-training for simulation experiment, we use the same encoder as for ST-DIM in the reduction phase.
For the decoder, we use the reverse architecture of the encoder. Precisely, we use a linear layer with 704 units followed by $4$ $1$D transpose convolution layers with output features $(128, 64, 32, 10)$, kernel sizes $(2, 3, 4, 4)$ respectively and all with stride $1$, that result in $10\times 20$ windows at the output.

In ST-DIM based pre-training, for all possible pairs in the batch, we
take features $c^3$ and $z$ after the $3$rd convolutional layer  and output layer respectively. We embed both $c^3$ and $z$ using $\psi$ and $\phi$ respectively to a $128$ dimensional vector to compute the score of a critic function $f_{global}$ or $f_{local}$. Using these scores, we compute the loss. The neural networks are trained using Adam optimizer~\cite{kingma2014adam}. The weights for encoder are initialized using orthogonal initialization. However, the LSTM and the final fully connected layer are initialized using Xavier~\cite{glorot2010understanding}.

As we are more interested in subjects for classification task, we feed each time series (ICA time courses) into the encoder in the form of a sequence of windows. The encoder encodes the windows of input data into latent representation. The latent representation of the entire time series is then concatenated and passed to a biLSTM with hidden dimension of size $200$. The output of biLSTM is then used as input to a feed forward network of two linear layers with $200$ and $2$ units to perform binary classification.

For experiments on the downstream tasks, a hold out is selected for testing and is never used through the training/validation phase. For each downstream task, the number of  subjects used for supervised training is gradually increased within a range to observe the effectiveness of pre-training in downstream task with varied number of  training subjects . For each experiment, 10 trials are performed to ensure random selection of training subjects and, in each case, the performance is evaluated on the hold out dataset (test data). 

For brain data, each of the models (\FPT{}, \UFPT{}, \NPT{}) yields the best results based on its respective gain value of Xavier~\cite{glorot2010understanding} initialization used for biLSTM. To find the best gain value for each model, $20$ values in the range $0-1$ are tried with an increment of $0.05$. For each value, $10$ experiments are performed and best value is chosen based on the results on validation dataset. Refer to Table~\ref{table:ParameterDetails} for more parametric details of the models.


\begin{table}[ht] 
\centering
\resizebox{\columnwidth}{!}{%
\begin{tabular}{l l l l }
\hline
\textbf{} & \textbf{Simulations} & \textbf{Real Data}\\
\hline
Training Batch Size & $16$ & min(32, no. of subjects) \\
Validation Batch Size & 200 & no. of subjects \\
Test Batch Size & 200 & no. of subjects \\
Initial Learning Rate & 3e-4 & 3e-4\\
Learning Rate Scheduler & None & Reduce LR on Plateau \\
Max Epochs & 500 & 3000 \\
\hline
\end{tabular}
}
\vspace{-1mm}
\caption{Parameter details of models used for experiments}
\label{table:ParameterDetails}

\end{table}

\subsection{Simulation}
To simulate the data, we generate multiple $10$-node graphs with $10 \times 10$ stable transition matrices. Using these we generate multivariate time series with autoregressive (VAR) and structural vector autoregressive (SVAR) models~\citep{lutkepohl2005new}. 

First, we generate $50$ VAR times series with size $10 \times 20000$. Then we split our dataset to $50\times10\times14000$ samples for training, $50\times10\times4000$ ---for validation and $50\times10\times2000$ --- for testing. Using these samples, 
We pre-train an encoder to learn consecutive windows (positive examples) from the same VAR time series. As mentioned in Section~\ref{Autoenc}, we also use autoencoder for pre-training the same encoder and show the effectiveness of ST-DIM to learn time-series dynamics in self-supervised manner. After pre-training, we use our pre-trained encoder for complete-time series classification. 

In the final downstream task, we classify the whole time-series whether it is generated by VAR or SVAR (undersampled VAR at rate 2). We create $400$ graphs with corresponding stable transition matrices and generate $2000 \times 10 \times 4000$ samples ($5$ for each) and split as $1600\times 10 \times 4000$ for training, $200\times 10 \times 4000$ for validation and $200\times 10 \times 4000$ for hold-out test. Here we also use $10 \times 20$ windows as a single time-point input.

First, we train our encoder to learn consecutive $10 \times 20$ windows from the VAR time series using InfoNCE based loss, and secondly, we train a supervised classifier based on windows. This window-based classification provides promising results. However, in solving similar real problems, we are more interested in subjects, i.e., entire time series, rather than a single-window for classification. Hence, we perform classification based on the whole time-series. In this setting, the entire time-series is encoded as a sequence of representations and fed through a biLSTM. Then, as described in Section \ref{Setup}, using the feed forward network on top of the last hidden state of the biLSTM, we map the representation to classification scores.
 
\begin{figure}[ht]
\includegraphics[width=\linewidth]{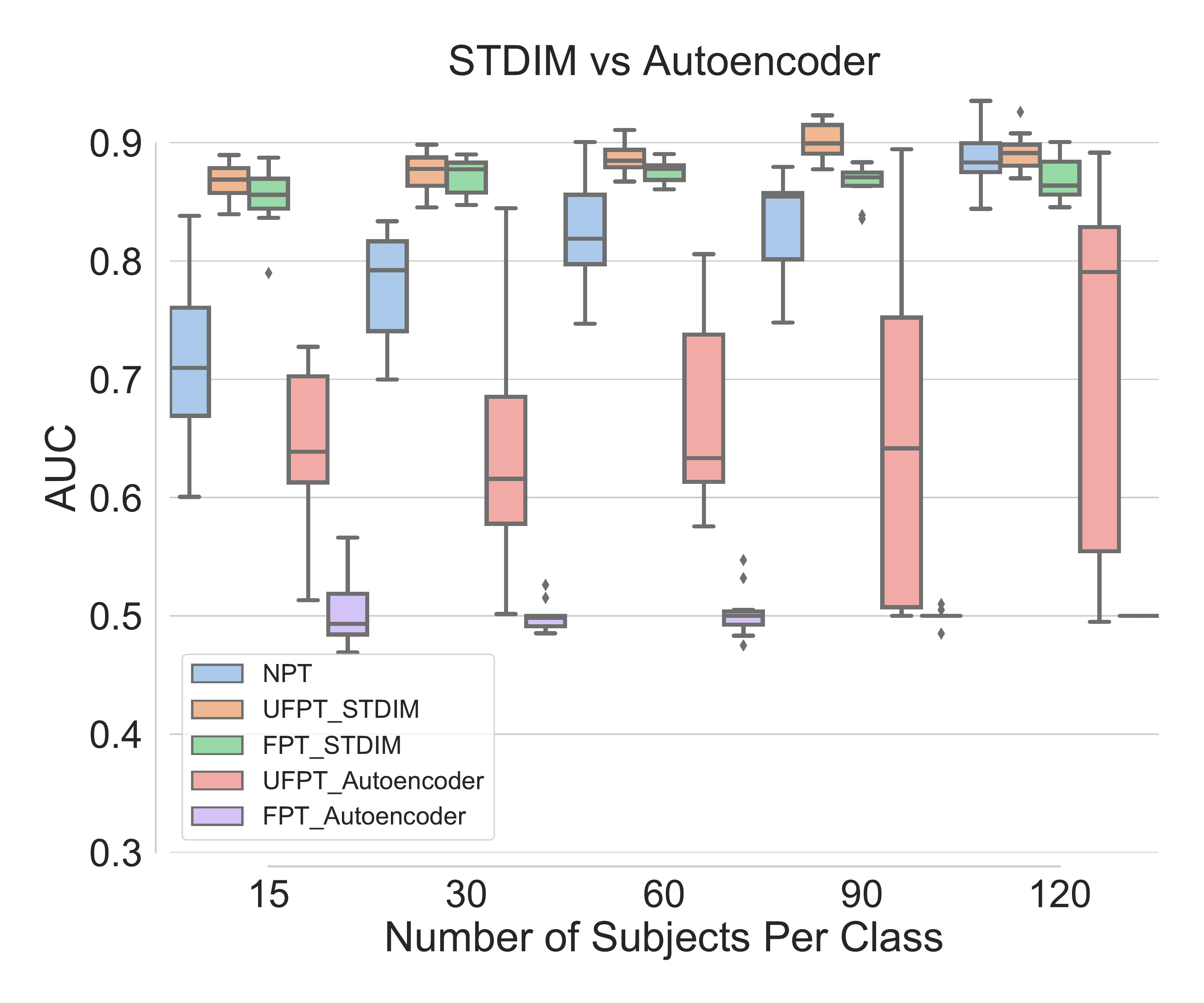}

  \caption{Area Under Curve (AUC) scores for VAR vs. SVAR time-series classification using ST-DIM and autoencoder based pre-training methods. ST-DIM based pre-training greatly improves the performance of downstream task with small datasets. On the other side, autoencoder based pre-training fails to learn dynamics and thus exhibits poor performance.}
  \label{fig:synth_test}
  \vspace{-1mm}
\end{figure}

 We observe that the ST-DIM based pre-trained models can easily be fine-tuned only with small amount of downstream data. In this classification task, our model can classify a randomly chosen time-series as a sample of VAR or SVAR. Note, with very few  samples, models based on the pre-trained encoder (\FPT{} and \UFPT{}) outperform supervised models. However, as the number of samples grows, the accuracy achieved with or without pre-training levels out. We also notice that autoencoder based self-supervised pre-training does not assist in VAR vs. SVAR classification. Consequently, we use only ST-DIM based pre-training for all the real data experiments. Refer to Figure \ref{fig:synth_test} for the results of simulation experiments.

\subsection{Keyword Detection}
To show the broad implications of unsupervised pre-training, we first apply it to a simple problem of keyword detection in audio files. We choose this problem as it has many practical applications (e.g., virtual assistants in smart phones, robots). 
We use LibriSpeech ASR corpus~\cite{librispeech} for pre-training and  Speech Commands Dataset~\cite{speechcommand} for supervised training. The audio files of both datasets are combined with a background noise of coffee shop collected from~\cite{backgroundnoise} to make pre-training and classification harder. 

\begin{figure}[ht]
  
  \centering
  \includegraphics[width=\linewidth]{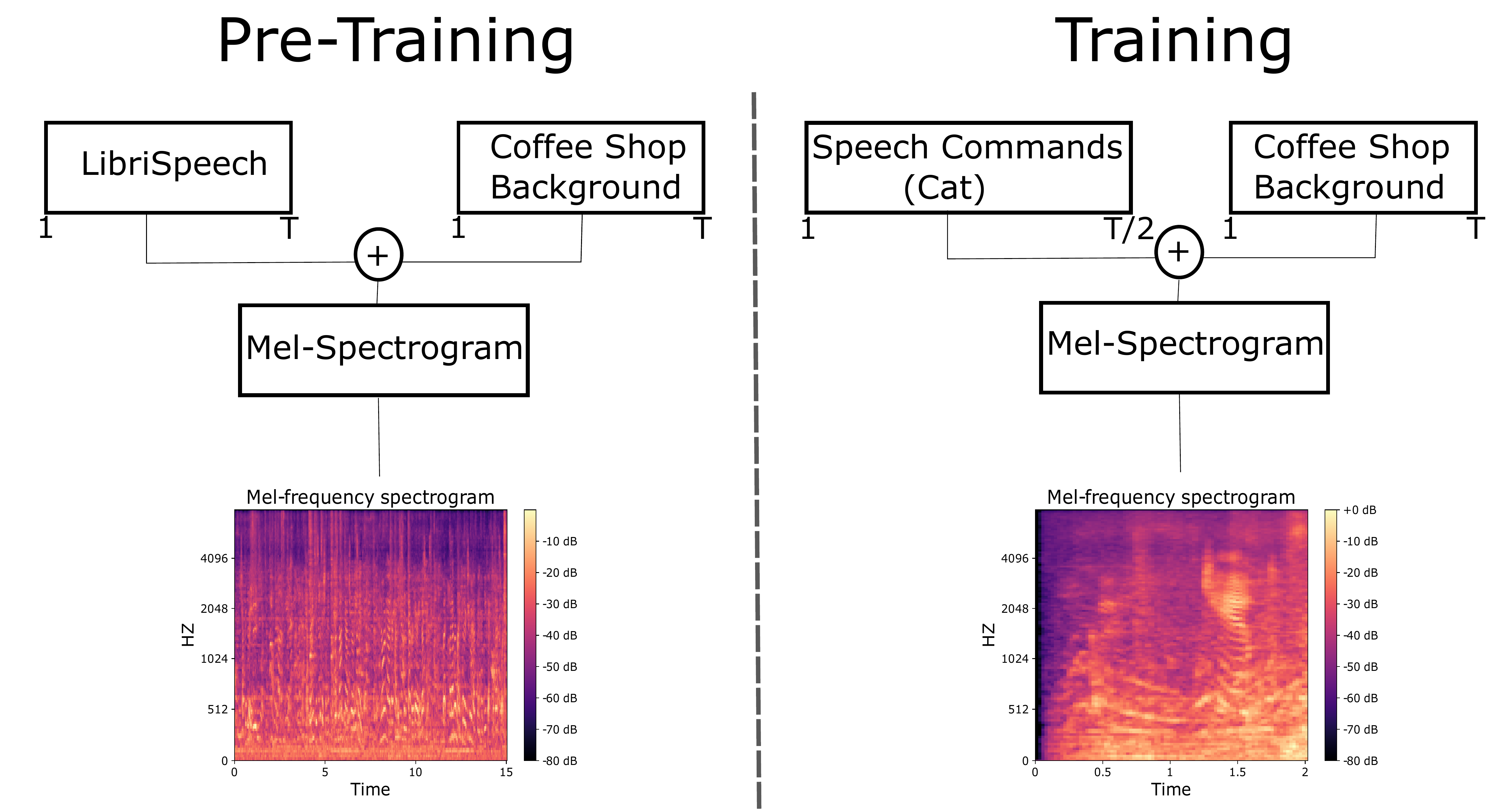} 
  \caption{\textbf{Left:} For pre-training, we combine audio files from LibriSpeech dataset with background noise of coffee shop. T is the length of audio files which ranges from $1$ to $20$ seconds. \textbf{Right:} For training, we superimpose a T/2 length audio of word "cat" padded with T/2 zeros onto a background noise of coffee shop of length T ($T=2$). For both pre-training and training, we calculate the mel-spectrogram of the combined audio files that results in a matrix of size \emph{components $\times$ time courses} for each audio file.}
  \label{speechpreprocessing}
  \vspace{-1mm}
\end{figure}

For pre-training, we calculate mel-spectrogram for all audio files from each subject and then concatenate together, resulting in a matrix of dimensions $C \times N$, where N, ranging anywhere from $10000$ to $60000$ depending on the length and number of audio files for each subject, is the time courses for the subject and $C = 128$ is the number of components in mel-spectrogram calculated at each time point. Figure~\ref{speechpreprocessing} left shows the pre-processing of one audio file from a subject used for pre-training.

Out of all the available subjects, we select 416 which have large number of time points $(N \geqslant 20k)$. We use 300 subjects for training and 116 for validation. For pre-training, we use the algorithm as described in section~\ref{STDIM} and achieve accuracy of $\thicksim 0.95$ on the validation dataset. 

For the downstream classification task, we collect samples of audios for the keyword "cat" from~\cite{speechcommand} which has 1515 audio files for the keyword. To create "cat" class examples, we superimpose each of the keyword audios having length of 1 second onto a 2 seconds long background noise at random location, resulting into a two seconds long audio consisting of background noise and keyword cat. Figure~\ref{speechpreprocessing} Right shows the pre-processing used in downstream classification for a "cat" example. To create "no cat" class examples, we use another 1515, 2 seconds long audio files containing only background noise of coffee shop. Thus, we use 3030 audio files for the downstream task ("cat"/"no cat" classification). For all of the 3030 audio files, we compute mel-spectrogram to convert each of them into a matrix of size \emph{components $\times$ time courses}. 

We perform 20 cross validations, each having a hold out of size 128 for test subjects. For each hold out, we conduct 5 different experiments varying number of samples (ranging from $15$ to $100$) from each class to be used for training. For each experiment, we perform 10 trials resulting into 200 trials to get accurate estimates of the performance. 

\begin{figure}[ht]
  \includegraphics[width=0.9\linewidth]{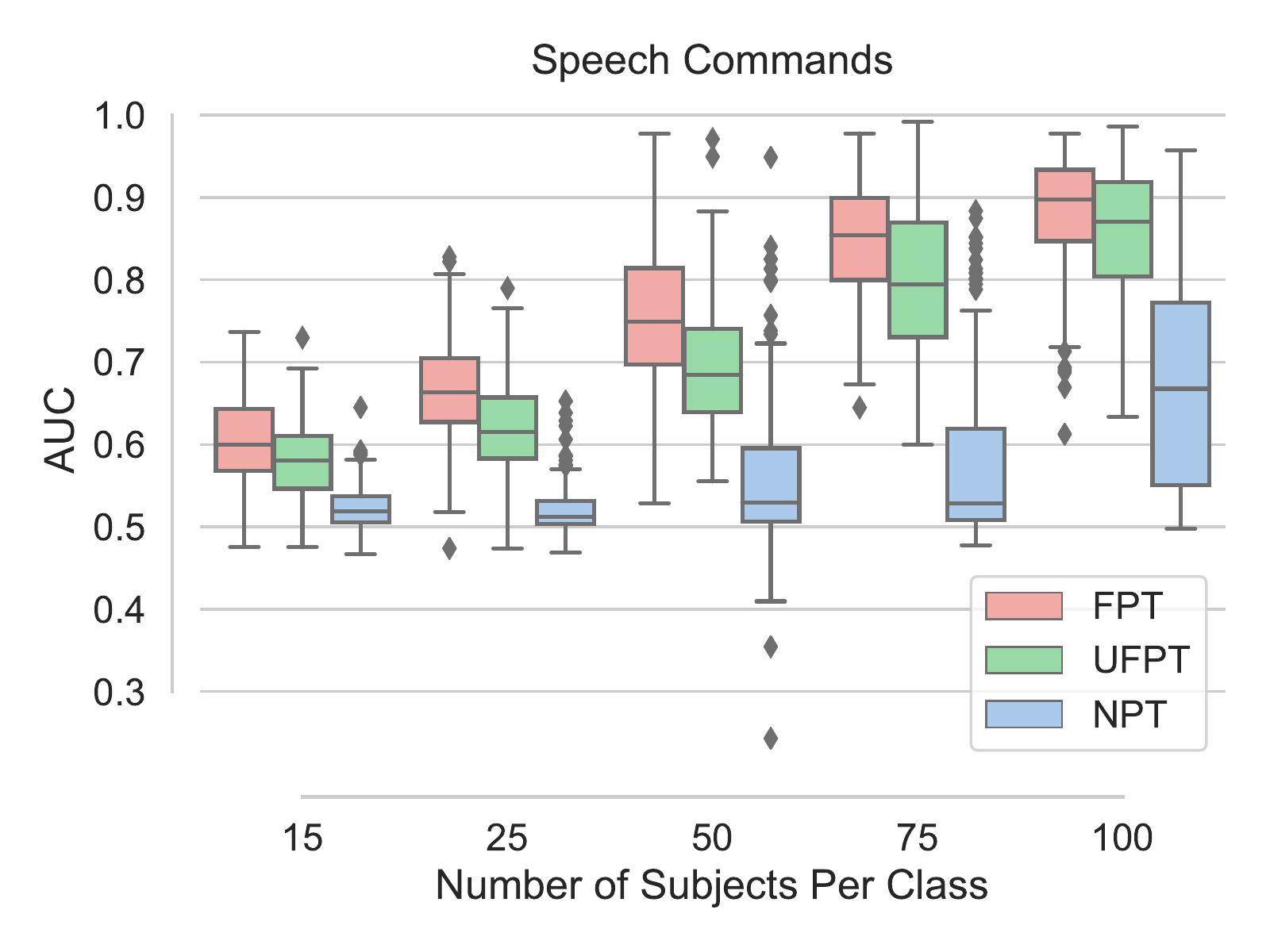}
  \centering
  \caption{AUC scores for all the models on keyword detection task. Notice that pre-trained models provide noticeably better performance than \NPT{} even with just $15$ subjects for training. The pre-trained models continue to improve significantly reaching more than $0.9$ with just 100 subjects, whereas \NPT\ fails to achieve even $0.8$ with the same number of training subjects.}
  \label{fig:real_test_AUC_Keyword}
  \vspace{-1mm}
\end{figure}

As observed in the simulation experiments, Figure~\ref{fig:real_test_AUC_Keyword} shows the importance of ST-DIM based unsupervised pre-training for keyword detection problem. Even with very few training subjects, \FPT{} and \UFPT{} outperform \NPT{} by a noticeable margin. As the number of samples increases, the performance of \NPT{} improves but still remains much lower than the pre-trained counterparts. 

\subsection{Brain Imaging}
\label{RealData}

\subsubsection{Datasets}
\begin{figure}[ht]
  \centering
  \includegraphics[width=0.7\linewidth]{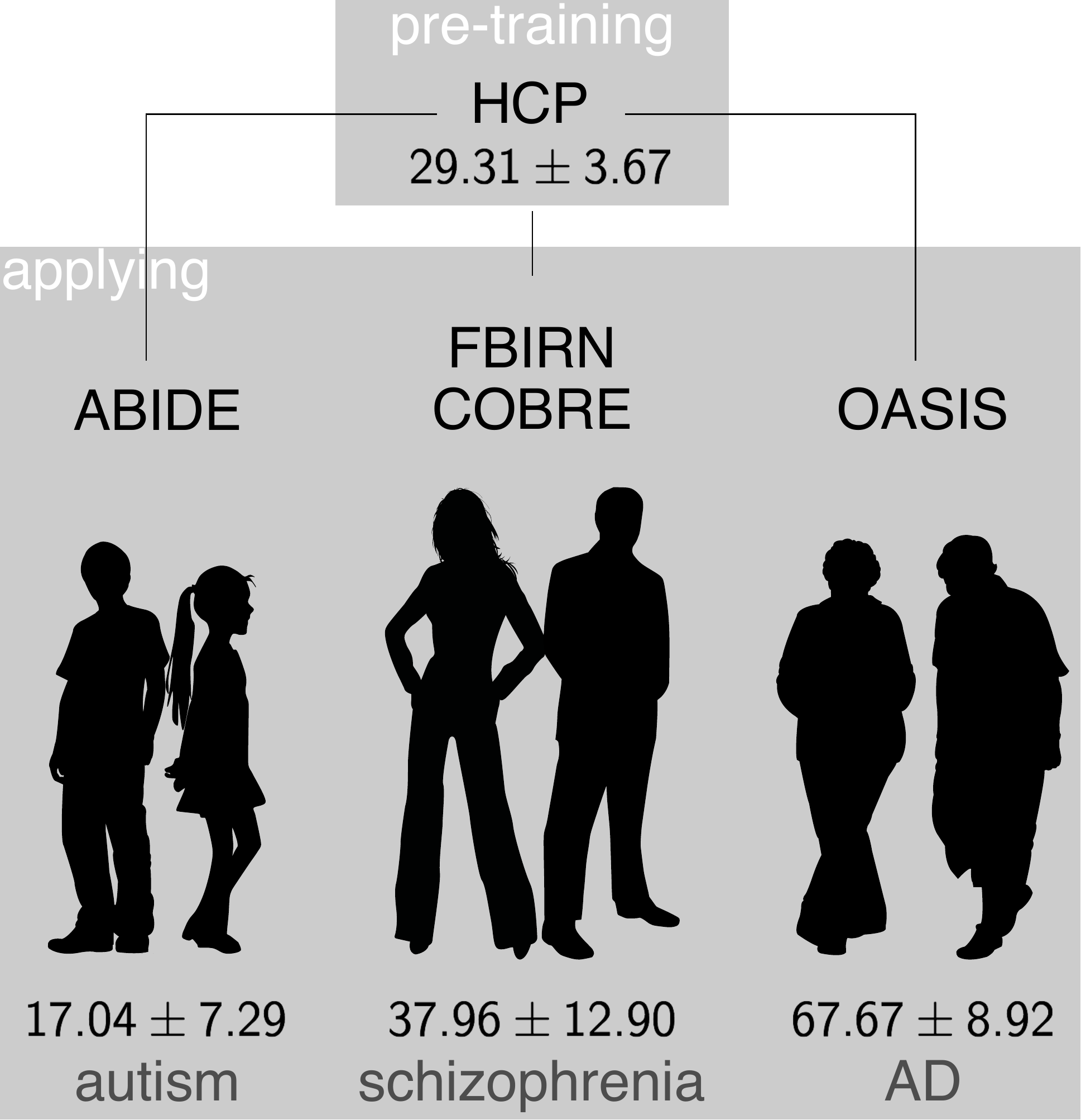} 
  \caption[Caption for LOF]{Datasets used for pre-training and classification tasks. Healthy controls from the HCP~\cite{van2013wu} are used for encoder pre-training guided by data dynamics alone\footnotemark[1]. The pre-trained encoder is then used in downstream classification tasks of $3$ different diseases, $4$ independently collected datasets, many of which contain data from a number of sites, and consist of populations with significant age difference. The age distributions in the datasets have the following means, medians and standard deviations: \textbf{HCP:} 29.31, 29.00, 3.667; \textbf{ABIDE:} 17.04, 15.40, 7.29; \textbf{COBRE:} 37.96, 37, 12.90; \textbf{FBIRN:} 37.87, 38, 11.25; \textbf{OASIS:} 67.67, 68, 8.92.}
  \label{braindataset}
  \vspace{-1mm}
\end{figure}

Next, we apply the same unsupervised pre-training method to brain imagining data. For encoder pre-training, we use HCP~\cite{van2013wu} consortium dataset and apply the pre-trained encoder for further downstream tasks. We apply the same pre-trained encoder for three different types of  diseases spanning four datasets to classify schizophrenia, autism and Alzheimer's diseases. We use resting fMRI data for all brain data experiments. Refer to Figure~\ref{braindataset} for the details of the datasets for disease classification. 

\footnotetext[1]{Human silhouettes are by Natasha Sinegina for
  Creazilla.com without modifications,
  \href{https://creativecommons.org/licenses/by/4.0/}{https://creativecommons.org/licenses/by/4.0/}}

Four datasets used in this study are collected from the FBIRN (Function Biomedical Informatics Research Network\footnote[2]{These data were downloaded from the Function BIRN Data Repository, Project Accession Number 2007-BDR-6UHZ1.})~\cite{keator2016function} project, from the COBRE (Center of Biomedical Research Excellence)~\cite{ccetin2014thalamus} project, from the release 1.0 of ABIDE (Autism Brain Imaging Data Exchange\footnote[3]{http://fcon\_1000.projects.nitrc.org/indi/abide/})~\cite{di2014autism} and from release 3.0 of OASIS (Open Access Series of Imaging Studies\footnote[4]{https://www.oasis-brains.org/})~\cite{rubin1998prospective}. 
Written informed consent was obtained from all participants of each dataset under protocols approved by the institutional review board (IRB).

\subsubsection{Preprocessing}
We preprocessed the fMRI data using statistical parametric mapping (SPM12, http://www.fil.ion.ucl.ac.uk/spm/) under MATLAB 2016 environment. A rigid body motion correction was performed using the toolbox in SPM to correct subject head motion, followed by the slice-timing correction to account for timing difference in slice acquisition. The fMRI data were subsequently warped into the standard Montreal Neurological Institute (MNI) space using an echo planar imaging (EPI) template and were slightly resampled to $3 \times 3 \times 3$ mm$^3$ isotropic voxels. The resampled fMRI images were finally smoothed using a Gaussian kernel with a full width at half maximum (FWHM) = $6$ mm. 
After the preprocessing. We included subjects in the analysis if the
subjects have head motion $\le 3^\circ$ and $\le 3$ mm, and with functional data providing near full brain successful normalization~\cite{fu2019altered}. 

\begin{figure}[ht]
  \includegraphics[width=0.9\linewidth]{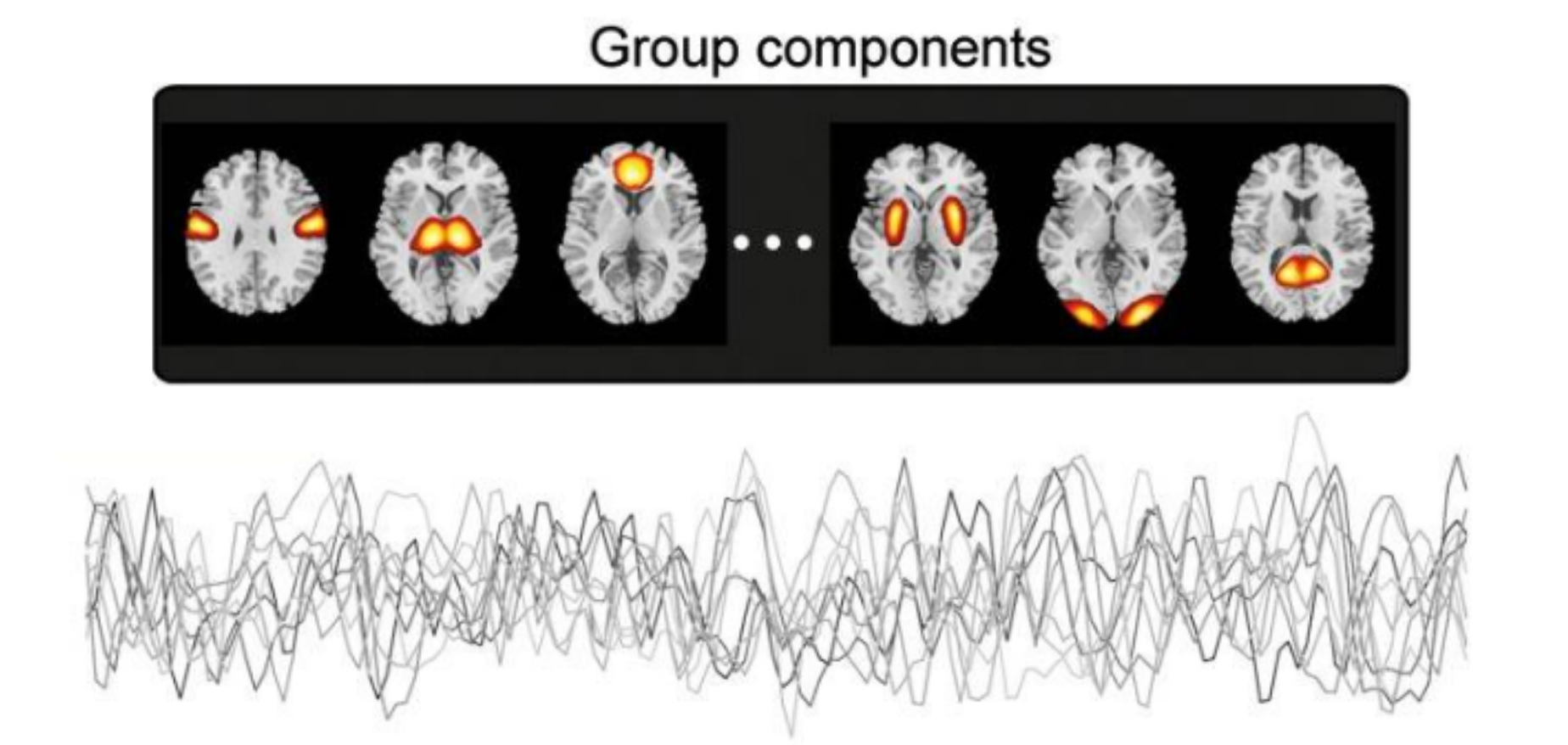}
  \centering
  \caption{ICA time courses are computed from the resting state fMRI data. Results contain statistically independent spatial maps (top) and their corresponding time courses.}
  \label{fig:brainsICA}
  \vspace{-1mm}
\end{figure}

For each dataset, $100$ ICA components as shown in Figure~\ref{fig:brainsICA} are acquired using the same procedure described in~\cite{fu2019altered}. However, only $53$ non-noise components as determined per slice (time point) are used in pre-training of encoder and  on downstream task. For experiments, including both pre-training and classification the fMRI sequence is divided into windows of $20$ time points. 

\subsubsection{Schizophrenia}
For schizophrenia classification, we conduct experiments on two different datasets, FBIRN~\cite{keator2016function} and COBRE~\cite{ccetin2014thalamus}. The datasets contain labeled Schizophrenia (SZ) and Healthy Control (HC) subjects.

\begin{figure}[ht]
  \includegraphics[width=0.9\linewidth]{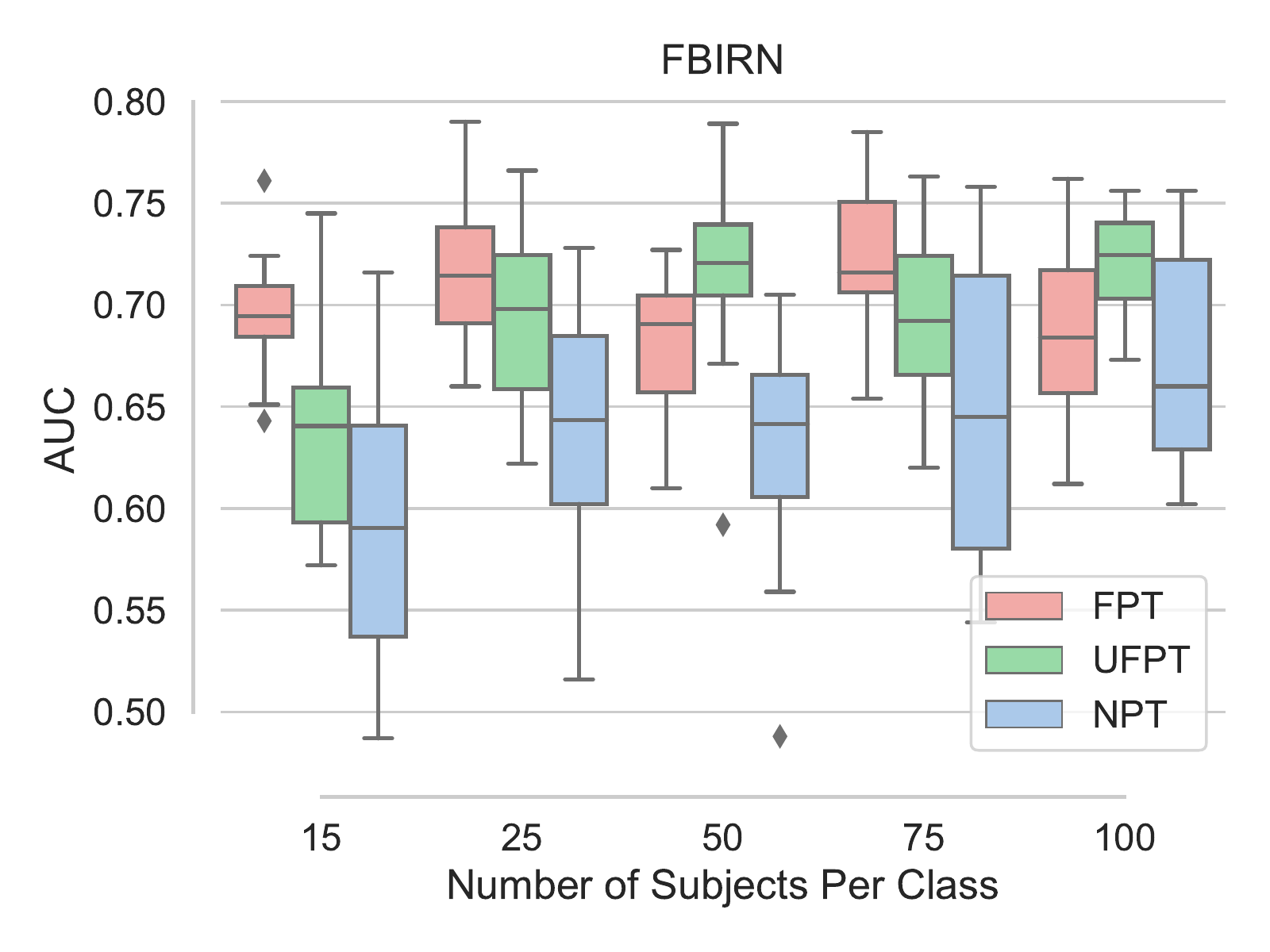}
  \centering
  \caption{AUC scores for all the three models (Refer to Figure \ref{fig:process}) on FBIRN dataset. It is noticeable that even with only $15$ subjects for supervised training, the median AUC scores of \FPT{} and \NPT{} differ by a large margin $(10\%)$.}
  \label{fig:real_test_AUC_}
  \vspace{-1mm}
\end{figure}

\noindent
{\bf FBIRN}
 The dataset has total $311$ subjects consisting of $150$ HC and $161$ affected with SZ. Each subject has $53$ non-noise components with $140$ time points. We use two hold-out sets with sizes $32$ and $64$ respectively for validation and test. The remaining data are used for supervised training. With $140$ time points, we create $53 \times 20$-sized windows with $50\%$ overlap along time dimension resulting in $13$ windows for each subject. The details of the results are shown in Figure~\ref{fig:real_test_AUC_}.
 
As we can see, the pre-trained models (\FPT\ and \UFPT) outperform \NPT. Even with very few subjects (15) for training, the difference between AUC scores obtained respectively with \FPT{} and \NPT{} is reasonably large ($\simeq 0.1$). 

\medskip 
\noindent
{\bf COBRE}
 The dataset has total $157$ subjects --- a collection of $68$ HC and $89$ affected with SZ. Like FBIRN, each subject has $53$ non-noise components in its ICA time courses with $140$ time points. We use two hold-out sets of size $32$ each respectively for validation and test. The remaining data are used for supervised training. With $140$ time points, we create $53 \times 20$-sized windows with no overlapping resulting in $7$ windows for each subject. Unlike FBIRN, it has been impossible to increase the number of subjects for downstream training due to insufficiency of data.
 
 \begin{figure}[ht]
  \includegraphics[width=0.9\linewidth]{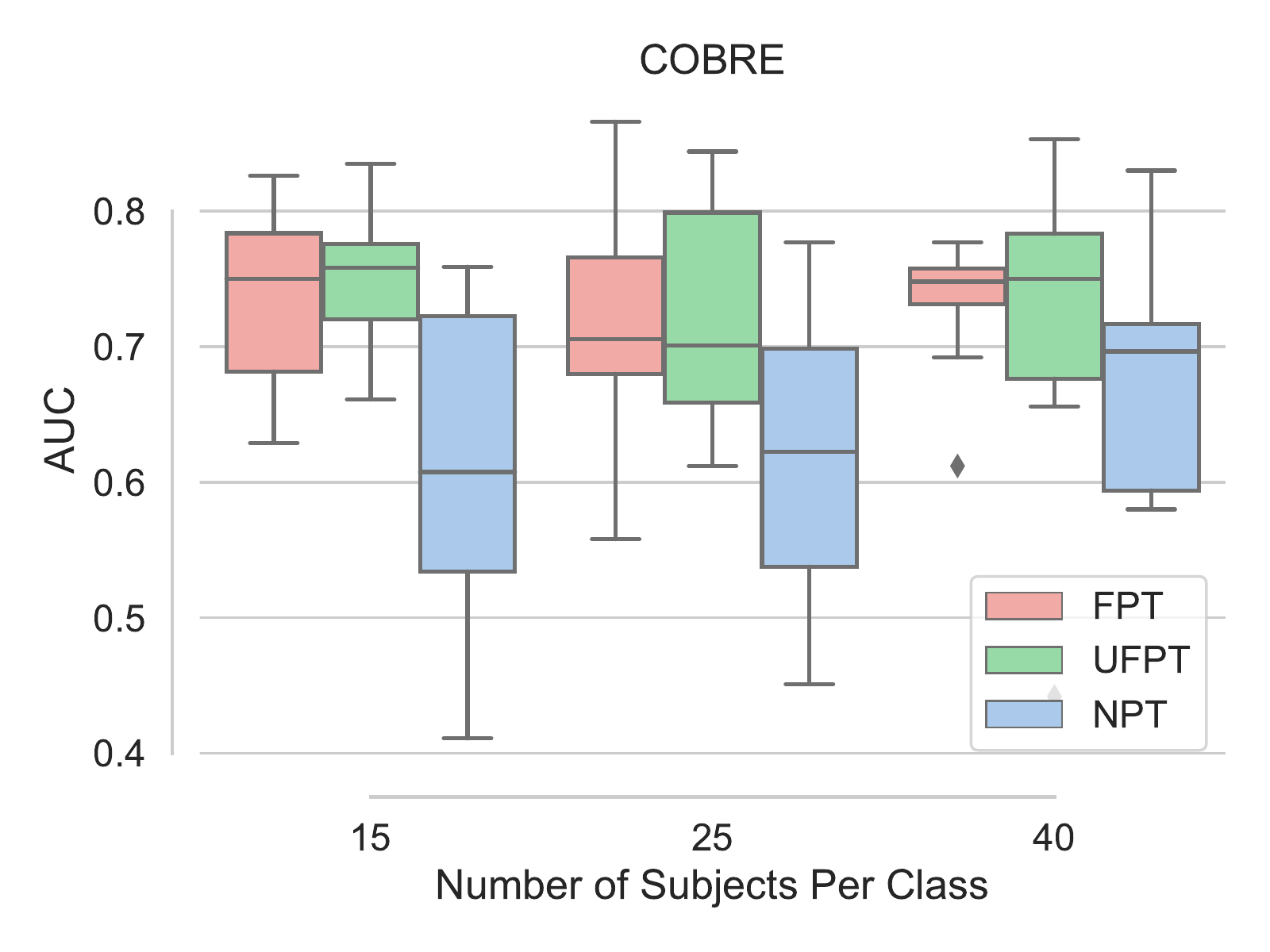}
  \centering
  \caption{AUC scores for all the three models on COBRE dataset. It is obvious that even with $15$ subjects for training, \FPT{} outperforms \NPT{} noticeably, that is, the difference between two median AUC scores is remarkable ($\simeq 0.15$).}
  \label{fig:real_test_AUC_COBRE}
  \vspace{-1mm}
\end{figure}

The results in Figure~\ref{fig:real_test_AUC_COBRE} adheres to the results of FBIRN. That is, with only $15$ training subjects, \FPT{} and \UFPT{} perform significantly better than \NPT{} having $\simeq 0.15$ difference in their median AUC scores. 

\subsubsection{Autism}
\begin{figure}[ht]
  \includegraphics[width=0.9\linewidth]{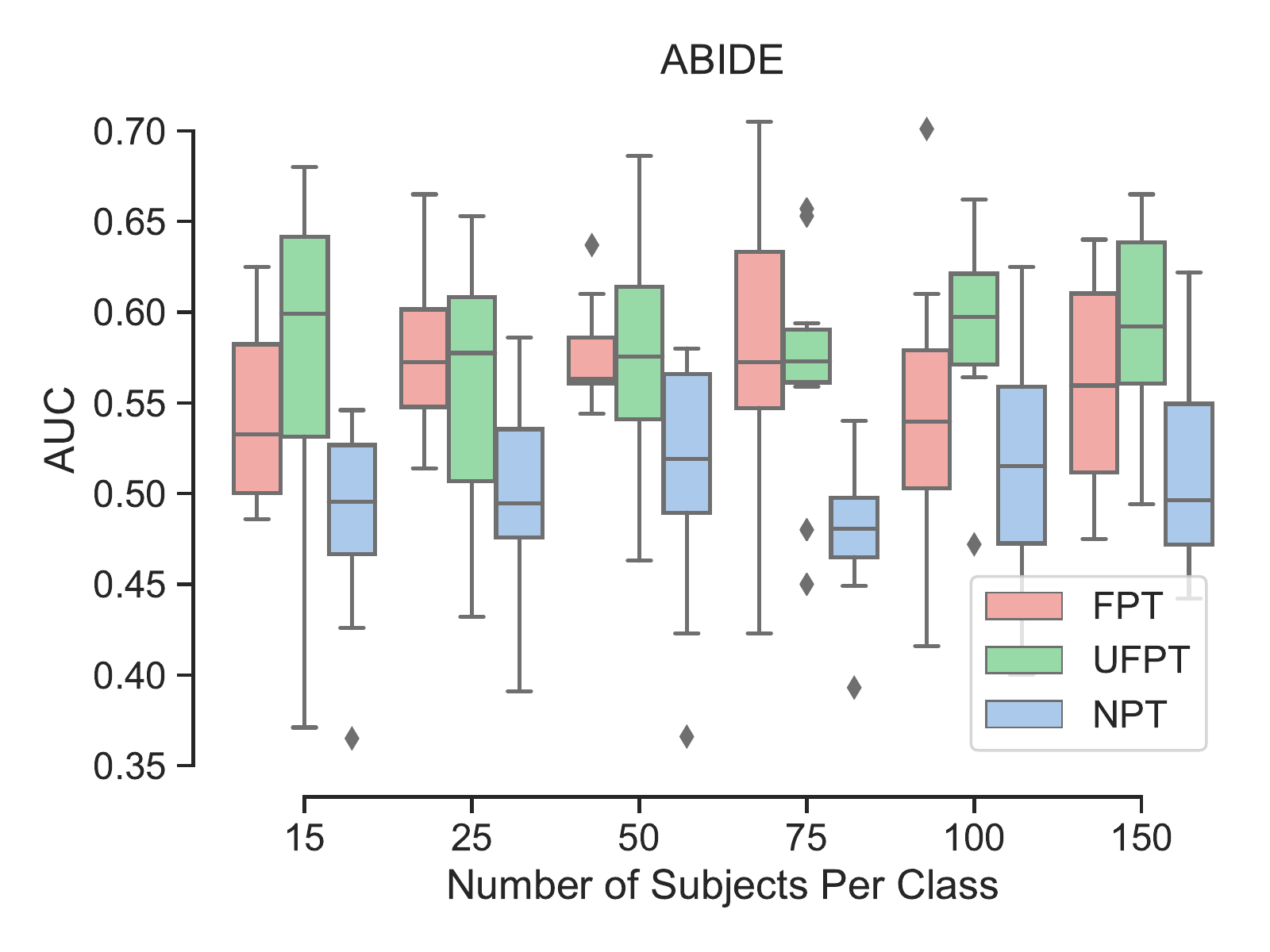}
  \centering
  \caption{AUC scores for all the three models on ABIDE dataset. Like experiments on FBIRN and COBRE, it is evident that the pre-trained models consistently perform better than \NPT{}.}
  \label{fig:real_test_AUC_ABIDE}
  \vspace{-1mm}
\end{figure}

The dataset ABIDE has total $569$ subjects, of which, $255$ are HC and $314$ are affected with autism. Like other datasets, each subject has $53$ non-noise components with $140$ time points. We use two hold-out sets of size $100$ each respectively for validation and test purpose. The remaining data are used for downstream training i.e., autism vs. HC classification. Like COBRE dataset, with $140$ time points, we create $53 \times 20$-sized windows with no overlapping resulting in $7$ windows for each subject. Refer to Figure~\ref{fig:real_test_AUC_ABIDE} for the details of the experimental results.

As seen in the figure, same pre-trained encoder performs reasonably better than \NPT{} for autism vs. HC classification and thus reinforces our hypothesis that unsupervised pre-training learns signal dynamics useful for downstream tasks. 
Note the difference between age ranges of ABIDE and HCP
datasets. The age range of ABIDE is much lower than that of HCP
  dataset used for pre-training. HCP dataset contains subjects of different ages with means 30.01 and 28.48, medians 30 and 28, and standard deviations 3.522 and 3.665 years respectively for female and male, whereas ABIDE dataset has overall mean 17.04, median 15.40 and standard deviation 7.29 years. Refer to Figure~\ref{braindataset} for the demographic information of all the datasets.   The dissimilarity in the age range is supposed to cause significant difference between these two datasets as the structure of brain and thought process of children is obviously different than adults. 

\subsubsection{Alzheimer's disease}

\begin{figure}[ht]
  \includegraphics[width=0.9\linewidth]{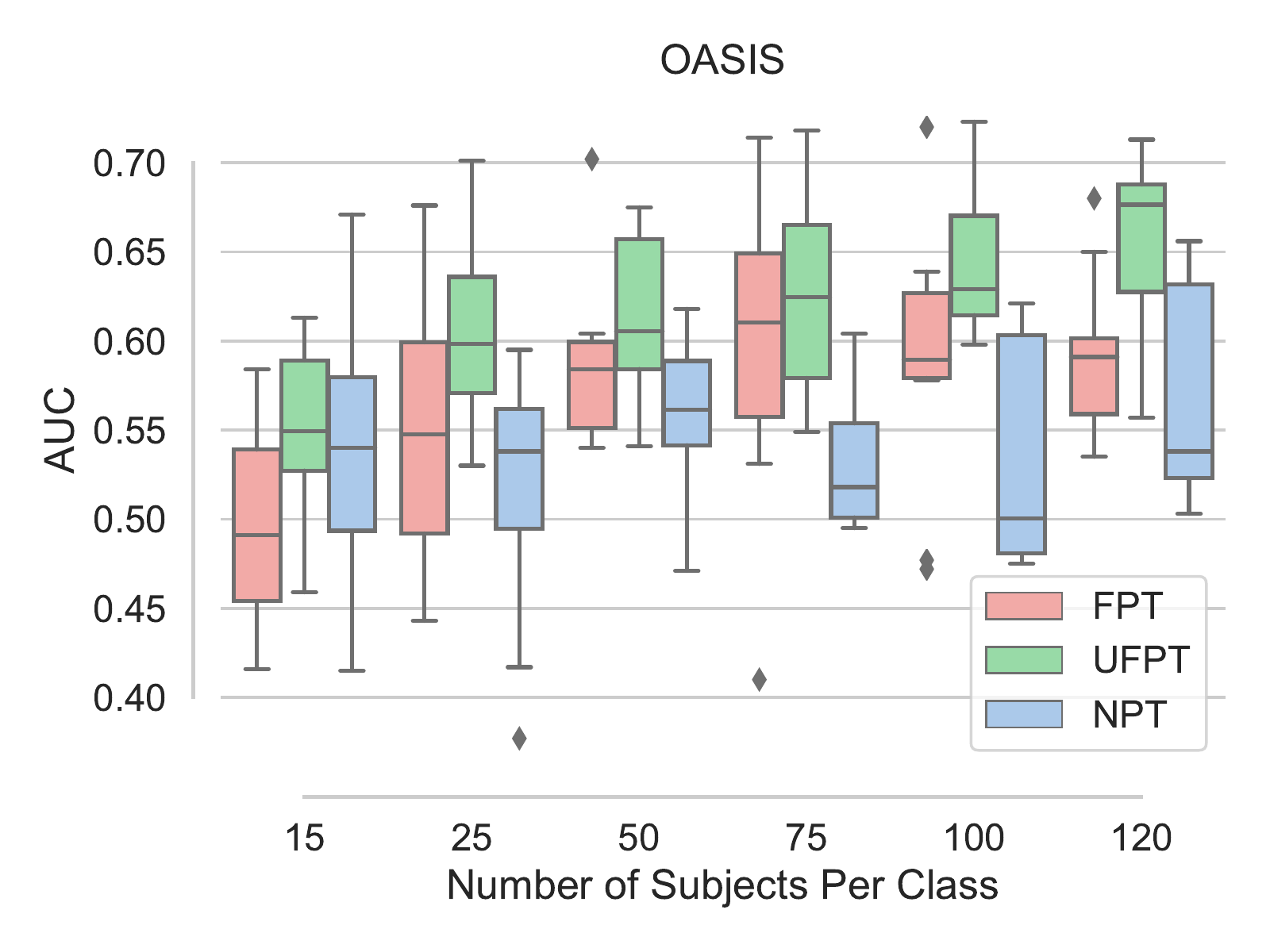}
  \centering
  \caption{AUC scores for all the models on OASIS dataset. As we continue increasing the number of subjects, the pre-trained models start learning and thus improve their respective scores. However, notice that the \NPT{} model even with $120$ subjects didn't significantly improve its predictability. }
  \label{fig:real_test_AUC_OASIS}
  \vspace{-1mm}
\end{figure}

The dataset OASIS~\cite{rubin1998prospective} has total $372$ subjects with equal number ($186$) of HC and AZ patients. We use two hold-out sets each of size $64$ respectively for validation and test purpose. The remaining are used for supervised training. Unlike other datasets described earlier, it has only $120$ time points though the number of non-noise componets is same ($53$) as other datasets. With $120$ time points, we use six $53 \times 20$-sized non-overlapping windows for each subject. Refer to Figure~\ref{fig:real_test_AUC_OASIS} for the details of the experiments. 

As we see, the AUC scores of all the three models for 15 subjects is $\thicksim 0.5$, which can be treated as merely random guess. However, as the number of subjects increases, pre-trained models gradually start performing better than \NPT{} which, in fact, even with $120$ subjects fails to learn from the data. We suspect that the reason why pre-trained models do not work for $15$ subjects is that the data set is much different than HCP. The big age gap between subjects of HCP and OASIS is a major difference and $15$ subjects are even not enough for pre-trained models.

\section{Conclusions and Future Work}

As we have demonstrated, self-supervised pre-training of a spatiotemporal encoder gives significant improvement on the downstream tasks in both keyword detection and brain imaging datasets. Pre-training on  fMRI of healthy subjects provides benefits that transfer across datasets, collection sites, and multiple disease classification with varying age gap. Learning dynamics of fMRI helps to improve classification results for schizophrenia, autism, Alzheimer's dieseases and speed up the convergence of the  algorithm on small datasets, that otherwise do not provide reliable generalizations. Although the utility of these results is highly promising by itself, we conjecture that direct application to spatio-temporal data will warrant benefits beyond improved classification accuracy in the future work. Working with ICA components is a smaller and thus easier to handle space that exhibits all dynamics of the signal, in future we will move beyond ICA pre-processing and work with fMRI data directly.  We expect model introspection to yield insight into the spatio-temporal biomarkers of schizophrenia.  In future work, we will use the same analogously pre-trained encoder on datasets with various other mental disorders such as MCI and bipolar. We are optimistic about the outcome because the proposed pre-training is oblivious of the downstream use and is done in a manner quite different from the classifier's work.  It may indeed be learning crucial information about dynamics that might contain important clues into the nature of mental disorders.

\section{Acknowledgement}
This study was  supported by startup funds  to SMP and in  part by NIH
grant R01EB020407.

Data for healthy subjects was provided [in part] by the Human Connectome Project, WU-Minn Consortium (Principal Investigators: David Van Essen and Kamil Ugurbil; 1U54MH091657) funded by the 16 NIH Institutes and Centers that support the NIH Blueprint for Neuroscience Research; and by the McDonnell Center for Systems Neuroscience at Washington University.

Data for Schizophrenia classification was used in this study were downloaded from the Function
BIRN Data Repository (http://bdr.birncommunity.org:8080/BDR/),
supported by grants to the Function BIRN (U24-RR021992) Testbed funded by the
National Center for Research Resources at the National Institutes of Health, U.S.A. and from the COllaborative Informatics and Neuroimaging
Suite Data Exchange tool (COINS; \href{http://coins.trendscenter.org}{http://coins.trendscenter.org}) and data collection was performed at the
Mind Research Network, and funded by a Center of Biomedical Research Excellence (COBRE)
grant 5P20RR021938/P20GM103472 from the NIH to Dr.Vince Calhoun.

Data for Alzheimer's was provided by OASIS-3: Principal Investigators: T. Benzinger, D. Marcus, J. Morris; NIH P50AG00561, P30NS09857781, P01AG026276, P01AG003991, R01AG043434, UL1TR000448, R01EB009352. AV-45 doses were provided by Avid Radiopharmaceuticals, a wholly owned subsidiary of Eli Lilly.

Autism data was provided by ABIDE. We acknowledge primary support for the work by Adriana Di Martino provided by the (NIMH K23MH087770) and the Leon Levy Foundation and primary support for the work by Michael P. Milham and the INDI team was provided by gifts from Joseph P. Healy and the Stavros Niarchos Foundation to the Child Mind Institute, as well as by an NIMH award to MPM (NIMH R03MH096321).

\bibliographystyle{unsrt}
\bibliography{references}

\begin{thebibliography}{10}

\bibitem{goldberg1992common}
David~P Goldberg and Peter Huxley.
\newblock {\em Common mental disorders: a bio-social model.}
\newblock Tavistock/Routledge, 1992.

\bibitem{calhoun2014chronnectome}
Vince~D Calhoun, Robyn Miller, Godfrey Pearlson, and Tulay Adal{\i}.
\newblock The chronnectome: time-varying connectivity networks as the next
  frontier in fmri data discovery.
\newblock {\em Neuron}, 84(2):262--274, 2014.

\bibitem{Khazaee2016}
Ali Khazaee, Ata Ebrahimzadeh, and Abbas Babajani-Feremi.
\newblock Application of advanced machine learning methods on resting-state
  {fMRI} network for identification of mild cognitive impairment and
  {A}lzheimer's disease.
\newblock {\em Brain Imaging and Behavior}, 10(3):799--817, Sep 2016.

\bibitem{yan2017discriminating}
Weizheng Yan, Sergey Plis, Vince~D Calhoun, Shengfeng Liu, Rongtao Jiang,
  Tian-Zi Jiang, and Jing Sui.
\newblock Discriminating schizophrenia from normal controls using resting state
  functional network connectivity: A deep neural network and layer-wise
  relevance propagation method.
\newblock In {\em 2017 IEEE 27th International Workshop on Machine Learning for
  Signal Processing (MLSP)}, pages 1--6. IEEE, 2017.

\bibitem{van2013wu}
David~C Van~Essen, Stephen~M Smith, Deanna~M Barch, Timothy~EJ Behrens, Essa
  Yacoub, Kamil Ugurbil, Wu-Minn~HCP Consortium, et~al.
\newblock The {WU-Minn} human connectome project: an overview.
\newblock {\em Neuroimage}, 80:62--79, 2013.

\bibitem{erhan2010does}
Dumitru Erhan, Yoshua Bengio, Aaron Courville, Pierre-Antoine Manzagol, Pascal
  Vincent, and Samy Bengio.
\newblock Why does unsupervised pre-training help deep learning?
\newblock {\em Journal of Machine Learning Research}, 11(Feb):625--660, 2010.

\bibitem{NIPS2007_3211}
Geoffrey~E Hinton and Russ~R Salakhutdinov.
\newblock Using deep belief nets to learn covariance kernels for gaussian
  processes.
\newblock In J.~C. Platt, D.~Koller, Y.~Singer, and S.~T. Roweis, editors, {\em
  Advances in Neural Information Processing Systems 20}, pages 1249--1256.
  Curran Associates, Inc., 2008.

\bibitem{vincent2008extracting}
Pascal Vincent, Hugo Larochelle, Yoshua Bengio, and Pierre-Antoine Manzagol.
\newblock Extracting and composing robust features with denoising autoencoders.
\newblock In {\em Proceedings of the 25th international conference on Machine
  learning}, pages 1096--1103. ACM, 2008.

\bibitem{krizhevsky2012imagenet}
Alex Krizhevsky, Ilya Sutskever, and Geoffrey~E Hinton.
\newblock Imagenet classification with deep convolutional neural networks.
\newblock In {\em Advances in neural information processing systems}, pages
  1097--1105, 2012.

\bibitem{radford2019language}
Alec Radford, Jeffrey Wu, Rewon Child, David Luan, Dario Amodei, and Ilya
  Sutskever.
\newblock Language models are unsupervised multitask learners.
\newblock {\em OpenAI Blog}, 1(8), 2019.

\bibitem{devlin2018bert}
Jacob Devlin, Ming-Wei Chang, Kenton Lee, and Kristina Toutanova.
\newblock Bert: Pre-training of deep bidirectional transformers for language
  understanding.
\newblock {\em arXiv preprint arXiv:1810.04805}, 2018.

\bibitem{mikolov2013distributed}
Tomas Mikolov, Ilya Sutskever, Kai Chen, Greg~S Corrado, and Jeff Dean.
\newblock Distributed representations of words and phrases and their
  compositionality.
\newblock In {\em Advances in neural information processing systems}, pages
  3111--3119, 2013.

\bibitem{gehring2013extracting}
Jonas Gehring, Yajie Miao, Florian Metze, and Alex Waibel.
\newblock Extracting deep bottleneck features using stacked auto-encoders.
\newblock In {\em 2013 IEEE international conference on acoustics, speech and
  signal processing}, pages 3377--3381. IEEE, 2013.

\bibitem{yu2010roles}
Dong Yu, Li~Deng, and George Dahl.
\newblock Roles of pre-training and fine-tuning in context-dependent dbn-hmms
  for real-world speech recognition.
\newblock In {\em Proc. NIPS Workshop on Deep Learning and Unsupervised Feature
  Learning}, 2010.

\bibitem{goodfellow2016}
Ian Goodfellow, Yoshua Bengio, and Aaron Courville.
\newblock {\em Deep learning}.
\newblock MIT press, 2016.

\bibitem{He_2019_ICCV}
Kaiming He, Ross Girshick, and Piotr Dollar.
\newblock Rethinking imagenet pre-training.
\newblock In {\em The IEEE International Conference on Computer Vision (ICCV)},
  October 2019.

\bibitem{oord2018representation}
Aaron van~den Oord, Yazhe Li, and Oriol Vinyals.
\newblock Representation learning with contrastive predictive coding.
\newblock {\em arXiv preprint arXiv:1807.03748}, 2018.

\bibitem{hjelm2018learning}
R~Devon Hjelm, Alex Fedorov, Samuel Lavoie-Marchildon, Karan Grewal, Phil
  Bachman, Adam Trischler, and Yoshua Bengio.
\newblock Learning deep representations by mutual information estimation and
  maximization.
\newblock {\em arXiv preprint arXiv:1808.06670}, 2018.

\bibitem{henaff2019data}
Olivier~J H{\'e}naff, Ali Razavi, Carl Doersch, SM~Eslami, and Aaron van~den
  Oord.
\newblock Data-efficient image recognition with contrastive predictive coding.
\newblock {\em arXiv preprint arXiv:1905.09272}, 2019.

\bibitem{bachman2019learning}
Philip Bachman, R~Devon Hjelm, and William Buchwalter.
\newblock Learning representations by maximizing mutual information across
  views.
\newblock {\em arXiv preprint arXiv:1906.00910}, 2019.

\bibitem{he2019momentum}
Kaiming He, Haoqi Fan, Yuxin Wu, Saining Xie, and Ross Girshick.
\newblock Momentum contrast for unsupervised visual representation learning,
  2019.

\bibitem{fedorov2019prediction}
Alex Fedorov, R~Devon Hjelm, Anees Abrol, Zening Fu, Yuhui Du, Sergey Plis, and
  Vince~D Calhoun.
\newblock Prediction of progression to {Alzheimers} disease with deep
  {InfoMax}.
\newblock {\em arXiv preprint arXiv:1904.10931}, 2019.

\bibitem{anand2019unsupervised}
Ankesh Anand, Evan Racah, Sherjil Ozair, Yoshua Bengio, Marc-Alexandre
  C{\^o}t{\'e}, and R~Devon Hjelm.
\newblock Unsupervised state representation learning in {A}tari.
\newblock {\em arXiv preprint arXiv:1906.08226}, 2019.

\bibitem{ravanelli2018learning}
Mirco Ravanelli and Yoshua Bengio.
\newblock Learning speaker representations with mutual information.
\newblock {\em arXiv preprint arXiv:1812.00271}, 2018.

\bibitem{imagenet_cvpr09}
J.~Deng, W.~Dong, R.~Socher, L.-J. Li, K.~Li, and L.~Fei-Fei.
\newblock {ImageNet: A Large-Scale Hierarchical Image Database}.
\newblock In {\em CVPR09}, 2009.

\bibitem{khosla2019machine}
Meenakshi Khosla, Keith Jamison, Gia~H Ngo, Amy Kuceyeski, and Mert~R Sabuncu.
\newblock Machine learning in resting-state {fMRI} analysis.
\newblock {\em Magnetic resonance imaging}, 2019.

\bibitem{frontiers2014}
Sergey~M Plis, Devon Hjelm, Ruslan Salakhutdinov, Elena~A Allen, Henry~Jeremy
  Bockholt, Jeffrey~D Long, Hans~J Johnson, Jane Paulsen, Jessica~A Turner, and
  Vince~D Calhoun.
\newblock Deep learning for neuroimaging: a validation study.
\newblock {\em Frontiers in Neuroscience}, 8(229), 2014.

\bibitem{calhoun2001method}
Vince~D Calhoun, Tulay Adali, Godfrey~D Pearlson, and JJ~Pekar.
\newblock A method for making group inferences from functional {MRI} data using
  independent component analysis.
\newblock {\em Human brain mapping}, 14(3):140--151, 2001.

\bibitem{eavani2013unsupervised}
Harini Eavani, Theodore~D Satterthwaite, Raquel~E Gur, Ruben~C Gur, and
  Christos Davatzikos.
\newblock Unsupervised learning of functional network dynamics in resting state
  fmri.
\newblock In {\em International Conference on Information Processing in Medical
  Imaging}, pages 426--437. Springer, 2013.

\bibitem{hjelm2014restricted}
R~Devon Hjelm, Vince~D Calhoun, Ruslan Salakhutdinov, Elena~A Allen, Tulay
  Adali, and Sergey~M Plis.
\newblock Restricted {B}oltzmann machines for neuroimaging: an application in
  identifying intrinsic networks.
\newblock {\em NeuroImage}, 96:245--260, 2014.

\bibitem{hjelm2018spatio}
R~Devon Hjelm, Eswar Damaraju, Kyunghyun Cho, Helmut Laufs, Sergey~M Plis, and
  Vince~D Calhoun.
\newblock Spatio-temporal dynamics of intrinsic networks in functional magnetic
  imaging data using recurrent neural networks.
\newblock {\em Frontiers in neuroscience}, 12:600, 2018.

\bibitem{khosla2019detecting}
Meenakshi Khosla, Keith Jamison, Amy Kuceyeski, and Mert~R Sabuncu.
\newblock Detecting abnormalities in resting-state dynamics: An unsupervised
  learning approach.
\newblock In {\em International Workshop on Machine Learning in Medical
  Imaging}, pages 301--309. Springer, 2019.

\bibitem{mensch2017learning}
Arthur Mensch, Julien Mairal, Danilo Bzdok, Bertrand Thirion, and Ga{\"e}l
  Varoquaux.
\newblock Learning neural representations of human cognition across many {fMRI}
  studies.
\newblock In {\em Advances in Neural Information Processing Systems}, pages
  5883--5893, 2017.

\bibitem{10.3389/fnins.2018.00491}
Hailong Li, Nehal~A. Parikh, and Lili He.
\newblock A novel transfer learning approach to enhance deep neural network
  classification of brain functional connectomes.
\newblock {\em Frontiers in Neuroscience}, 12:491, 2018.

\bibitem{thomas2019deep}
Armin~W Thomas, Klaus-Robert M{\"u}ller, and Wojciech Samek.
\newblock Deep transfer learning for whole-brain {fMRI} analyses.
\newblock {\em arXiv preprint arXiv:1907.01953}, 2019.

\bibitem{ulloa2018improving}
Alvaro Ulloa, Sergey Plis, and Vince Calhoun.
\newblock Improving classification rate of schizophrenia using a multimodal
  multi-layer perceptron model with structural and functional {MR}.
\newblock {\em arXiv preprint arXiv:1804.04591}, 2018.

\bibitem{infonce}
Aaron van~den Oord, Yazhe Li, and Oriol Vinyals.
\newblock Representation learning with contrastive predictive coding.
\newblock {\em arXiv preprint arXiv:1807.03748}, 2018.

\bibitem{tschannen2019mutual}
Michael Tschannen, Josip Djolonga, Paul~K Rubenstein, Sylvain Gelly, and Mario
  Lucic.
\newblock On mutual information maximization for representation learning.
\newblock {\em arXiv preprint arXiv:1907.13625}, 2019.

\bibitem{autoenc1}
Mark~A Kramer.
\newblock Nonlinear principal component analysis using autoassociative neural
  networks.
\newblock {\em AIChE journal}, 37(2):233--243, 1991.

\bibitem{schuster1997bidirectional}
Mike Schuster and Kuldip~K Paliwal.
\newblock Bidirectional recurrent neural networks.
\newblock {\em IEEE Transactions on Signal Processing}, 45(11):2673--2681,
  1997.

\bibitem{glorot2011deep}
Xavier Glorot, Antoine Bordes, and Yoshua Bengio.
\newblock Deep sparse rectifier neural networks.
\newblock In {\em Proceedings of the fourteenth international conference on
  artificial intelligence and statistics}, pages 315--323, 2011.

\bibitem{kingma2014adam}
Diederik~P Kingma and Jimmy Ba.
\newblock Adam: A method for stochastic optimization.
\newblock {\em arXiv preprint arXiv:1412.6980}, 2014.

\bibitem{glorot2010understanding}
Xavier Glorot and Yoshua Bengio.
\newblock Understanding the difficulty of training deep feedforward neural
  networks.
\newblock In {\em Proceedings of the thirteenth international conference on
  artificial intelligence and statistics}, pages 249--256, 2010.

\bibitem{lutkepohl2005new}
Helmut L{\"u}tkepohl.
\newblock {\em New introduction to multiple time series analysis}.
\newblock Springer Science \& Business Media, 2005.

\bibitem{librispeech}
V.~Panayotov, G.~Chen, D.~Povey, and S.~Khudanpur.
\newblock Librispeech: an {ASR} corpus based on public domain audio books.
\newblock {\em ICASSP}, pages 5206--5210, 2015.

\bibitem{speechcommand}
Pete Warden.
\newblock Speech commands: A dataset for limited-vocabulary speech recognition.
\newblock 2018.

\bibitem{backgroundnoise}
Duff~The Psych.
\newblock One hour of hq coffee shop background noise.
\newblock 2014.

\bibitem{keator2016function}
David~B Keator, Theo~GM van Erp, Jessica~A Turner, Gary~H Glover, Bryon~A
  Mueller, Thomas~T Liu, James~T Voyvodic, Jerod Rasmussen, Vince~D Calhoun,
  Hyo~Jong Lee, et~al.
\newblock The function biomedical informatics research network data repository.
\newblock {\em Neuroimage}, 124:1074--1079, 2016.

\bibitem{ccetin2014thalamus}
Mustafa~S {\c{C}}etin, Fletcher Christensen, Christopher~C Abbott, Julia~M
  Stephen, Andrew~R Mayer, Jos{\'e}~M Ca{\~n}ive, Juan~R Bustillo, Godfrey~D
  Pearlson, and Vince~D Calhoun.
\newblock Thalamus and posterior temporal lobe show greater inter-network
  connectivity at rest and across sensory paradigms in schizophrenia.
\newblock {\em Neuroimage}, 97:117--126, 2014.

\bibitem{di2014autism}
Adriana Di~Martino, Chao-Gan Yan, Qingyang Li, Erin Denio, Francisco~X
  Castellanos, Kaat Alaerts, Jeffrey~S Anderson, Michal Assaf, Susan~Y
  Bookheimer, Mirella Dapretto, et~al.
\newblock The autism brain imaging data exchange: towards a large-scale
  evaluation of the intrinsic brain architecture in autism.
\newblock {\em Molecular psychiatry}, 19(6):659, 2014.

\bibitem{rubin1998prospective}
Eugene~H Rubin, Martha Storandt, J~Philip Miller, Dorothy~A Kinscherf,
  Elizabeth~A Grant, John~C Morris, and Leonard Berg.
\newblock A prospective study of cognitive function and onset of dementia in
  cognitively healthy elders.
\newblock {\em Archives of neurology}, 55(3):395--401, 1998.

\bibitem{fu2019altered}
Zening Fu, Arvind Caprihan, Jiayu Chen, Yuhui Du, John~C Adair, Jing Sui,
  Gary~A Rosenberg, and Vince~D Calhoun.
\newblock Altered static and dynamic functional network connectivity in
  alzheimer's disease and subcortical ischemic vascular disease: shared and
  specific brain connectivity abnormalities.
\newblock {\em Human Brain Mapping}, 2019.

\end{thebibliography}

\end{document}